\documentclass[runningheads]{llncs}
\usepackage[T1]{fontenc}
\usepackage[noend]{algpseudocode}
\usepackage{algorithm}
\usepackage{graphicx}
\graphicspath{{figures/}}
\usepackage{xspace}
\usepackage{standalone}
\usepackage{tikz}
\usepackage{marvosym} 
\usepackage{subcaption}
\usepackage{multirow}
\usepackage{booktabs}
\usepackage{longtable}
\usepackage{array} 
\usepackage{enumitem}
\newlist{tabenum}{enumerate}{1}
\setlist[tabenum]{wide=0pt, 
                  nosep, 
                  leftmargin= * ,
                  label*=\arabic*.,
                  after=\vspace{-\baselineskip},
                  before=\vspace{-0.6\baselineskip}}

\usepackage{researchpack}
\usepackage{hyperref}
\usepackage[capitalise,nameinlink]{cleveref}
\usepackage{wrapfig}
\newcommand{\ilog}{iLog\xspace}
\newcommand{\skel}{\textsc{skel}\xspace}

\newcommand{\noisyy}{\ensuremath{\tilde{y}}\xspace}
\newcommand{\predy}{\ensuremath{\hat{y}}\xspace}

\newcommand{\fone}{\ensuremath{F_1}}
\newcommand{\gpnever}{$\textsc{gp}_\text{never}$\xspace}

\begin{document}
\title{Help the machine to help you:\\an evaluation in the wild of egocentric data cleaning via skeptical learning}

\titlerunning{Help the machine to help you}
%
\author{Andrea Bontempelli\inst{1}\thanks{Corresponding author.}\orcidID{0000-0001-7037-5797} \and
Matteo Busso\inst{1}\orcidID{0000-0002-3788-0203} \and
Leonardo Javier Malcotti\inst{1}\orcidID{0009-0000-0589-6393} \and
Fausto Giunchiglia\inst{1}\orcidID{0000-0002-5903-6150}}
\authorrunning{A. Bontempelli et al.}
%
\institute{University of Trento, Trento, Italy\\
\email{name.surname@unitn.it}}
\maketitle              
\begin{abstract}
Any digital personal assistant, whether used to support task performance, answer questions, or manage work and daily life—including fitness schedules—requires high-quality annotations to function properly. However, user annotations, whether actively produced or inferred from context (e.g., data from smartphone sensors), are often subject to errors and noise.
Previous research on Skeptical Learning (\skel) addressed the issue of noisy labels by comparing offline active annotations with passive data, allowing for an evaluation of annotation accuracy. However, this evaluation did not include confirmation from end-users, the best judges of their own context. 
In this study, we evaluate \skel's performance in real-world conditions with actual users who can refine the input labels based on their current perspectives and needs. The study involves university students using the iLog mobile application on their devices over a period of four weeks. The results highlight the challenges of finding the right balance between user effort and data quality, as well as the potential benefits of using \skel, which include reduced annotation effort and improved quality of collected data.

\end{abstract}
\section{Introduction}
\label{sec:exp:intro}
Any personal assistant requires knowledge of the user's context to operate effectively. For instance, a navigation app needs to know the user's current location and their intended destination to provide the best route. Similarly, a smart home assistant must recognize the user's voice and identify the connected smart devices, and even a simple calendar application requires information about upcoming events to send notifications. This contextual information can often be inferred without direct user interaction, such as through a smartphone's GPS signal indicating location. However, it frequently necessitates direct input from the user, who possesses the most comprehensive understanding of their own context~\cite{bontempelli2022lifelong}. Thus, personal context can be defined as a ``\textit{theory of the world which encodes an individual's subjective perspective about it}'' \cite{faustogiunchigliaContextualReasoning1993}. Following the approaches identified by ~\cite{giunchigliaPersonalContextModelling2017}, context can be modeled across four dimensions: \textit{location} (where the person is), \textit{activity} (what they are doing), \textit{social context} (who the person is with), and \textit{object context} (which objects are with the person). This information, which may come as feedback, vocal or written notes from the user, enhances the assistant's comprehension of both context and the goals it aims to achieve.

Past experiences in AI~\cite{diversityOne2025} and in social science in general, particularly through methods like Experience Sampling Method (ESM)~\cite{csikszentmihalyi2014validity,myin2022esm} or Ecological Momentary Assessment (EMA), demonstrate how this type of information can be collected continuously and frequently over extended periods. This collection process can combine user annotations with data from sensors, thus allowing for the analysis of habits, routines, and social practices. Frequent context questions lead to granular data, enabling the observation of more detailed behaviors. However, this approach is susceptible to errors that can compromise data quality (see, e.g., \cite{biemer2017total,weisberg2009total,myin2022esm}), both from user annotations and sensor information, which may end up in producing noisy labels. To tackle the problem of noisy labels, Skeptical Learning detects potentially suspicious responses and prompts the user to revise them when necessary ~\cite{zeni2019fixing,bontempelliLearningWildIncremental2021}. Prior studies have evaluated \skel using both synthetic and real-world datasets, where an oracle simulates user responses to assess the algorithm’s performance. The oracle reflects an objective point of view, relying on methods such as deriving location from GPS coordinates. These evaluations often made assumptions about users and data to lessen the impact of external factors. For instance, they typically require that examples arrive in a specific order, and assume that users respond immediately. However, these assumptions overlook the complexities of real life; users may leave the experiment, or sensor data may be unavailable if the user disables the sensor (like GPS) or if it is not present on their device.

In this work, we evaluate \skel with real users in the wild. We do this, as proposed in \cite{bontempelliLearningWildIncremental2021}, in a longitudinal study based on the use of the participants' smartphones. 
In longitudinal studies, the benefit of \skel is twofold. First, the ML model recognizes the participant's context from the sensor data to automatically answer questions about their context, thus reducing the effort required for answering. Second, \skel improves the answer quality, therefore allowing to learn a model with higher predictive performance and improving the accuracy of the answers.
The model is trained on a per-user basis, using only the user data, ensuring an egocentric recognition bound to the point of view of a single person, as done in the computer vision domain~\cite{erculiani2020continual}. This is because the context is user-specific and captures the subjective perspective. For instance, both the student and the professor are in the same room, but they are performing different activities, i.e., listening and speaking, respectively. The user is thus empowered to clean his/her own contextual annotation.
We designed and ran a study with university students from the University of Trento, aiming to identify their locations using \skel. 
\footnote{This experiment involves human subjects and has been approved by the Research Ethics Committee of the University of Trento (protocol n. 2023-006).}
The ultimate goal of \skel is to reduce the number of questions sent to the user by asking the model to answer the questions about the user's position. Our main contributions are as follows:
\begin{itemize}
    \item We leverage \skel as a solution to annotation errors in longitudinal studies;
    \item We design a study and operationalize \skel inside an existing data collection platform to evaluate it in the wild with real users and report preliminary results.
\end{itemize}
The structure of the paper is as follows. \cref{sec:related_works} presents how data quality is addressed in machine learning and in longitudinal studies, and \cref{sec:bck} briefly introduce \skel. Then, we detail the experiment design in \cref{sec:exp:design} and the results in \cref{sec:exp:analysis}. \cref{sec:conclusion} reports final remarks.

\section{Related Work}\label{sec:related_works}

\subsection{Data quality in machine learning}

In data-driven disciplines such as machine learning, the quality of data is fundamental for reliable model performance. Data may fail to capture the diversity of users and contexts in which a model will operate, while classification labels are often noisy due to annotation errors, leading to performance degradation. Several strategies exist to mitigate these issues, such as developing robust models, cleaning the data~\cite{frenay2014classification}, or collecting additional datasets. However, large-scale data acquisition is frequently infeasible due to cost and logistical constraints, creating a trade-off between dataset size and data quality, as well as between the use of large and small models \cite{varoquauxHypeSustainabilityPrice2024}.

To address these challenges, data-centric AI has emerged as a paradigm that prioritizes iterative improvement of data quality over simply scaling models \cite{liangAdvancesChallengesOpportunities2022}. Within this paradigm, our work emphasizes the \textit{human-in-control} principle, where users themselves are empowered to actively clean and refine their data.

\subsection{Data collection in the social sciences}

In the social sciences, researchers often collect data on human behavior using participants’ personal mobile devices. This approach integrates the experience sampling method (ESM) \cite{csikszentmihalyi2014validity,myin2022esm}—a longitudinal design in which participants repeatedly report on feelings, contexts, and behaviors—with passive sensor data \cite{van2017experience}. Questionnaires are typically delivered via smartphones and smartwatches, allowing high-frequency data collection while simultaneously capturing sensor-based contextual information such as location, activity, and social interactions \cite{diversityOne2025}. This combination enables fine-grained insights into daily routines and practices.

Extensive literature highlights how the quality of measurements and the dynamics of participant–researcher interaction influences the reliability of collected data \cite{corbetta2003social,weisberg2009total}.

\subsection{Reliability issues in collected data}

Self-reported data, while widely used, are prone to several biases. The asymmetric relationship between researcher and participant can give rise to phenomena such as the \textit{Hawthorne effect} \cite{hart1943hawthorne,mccarney2007hawthorne}, where participants alter their behavior due to awareness of observation. Similarly, \textit{social desirability} bias leads respondents to provide answers perceived as favorable \cite{corbetta2003social}, while \textit{non-attitudes} occur when participants respond at random due to lack of knowledge or understanding of the question.
Another critical issue is \textit{respondent burden} \cite{rolstad2011burden}, which reflects the perceived effort of participation, including time investment, difficulty, and emotional cost. In intensive longitudinal studies, this burden can reduce data quality through missing or careless responses. Several works have focused on the reasons linked to the respondent burden \cite{eisele2022effects} and ways to reduce its impact, such as through more immediate and short interactions, also called $\mu$EMA \cite{ponnada2025ema}.

Although less studied, sensor-based data collection also raises reliability concerns~\cite{busso24methodology}. Sensing technologies, such as smartphone-based measurements, are often seen as more “objective” alternatives to self-reports, yet they introduce specific challenges. Participants may disable notifications, turn off devices, or deactivate sensors (e.g., GPS, Bluetooth) for privacy or convenience, while technical issues—such as server crashes, notification failures, operating system incompatibilities, or missing hardware—can further compromise data quality.
Thus, both annotations and sensors data are vulnerable to reliability issues, though through different mechanisms. Addressing these challenges is essential to balance participant burden, data quality, and technological feasibility.

\section{Skeptical Learning}\label{sec:bck}

The benefit of \skel in the scenario described above is twofold. First, the machine learning model can precompile or autonomously answer these questions when certain about its prediction, reducing the effort of the participants and maintaining a higher level of granularity in the answer compared to $\mu$EMA. Second, the quality of the answers can be improved both to obtain better data for the researchers and to train a more accurate model.
We first introduce the main algorithm and then describe how we have applied it in the study described in this paper.

\subsection{\skel}

We tackle interactive classification in the wild, namely under label noise. The machine receives a sequence of examples $\vx_t$, for $t=1,2,\ldots$, where $\vx_t \in \bbR^d$, i.e., the vector of sensor readings, and outputs $\predy_t \in \bbY$, i.e., the user's location. The machine can query the user to report the ground-truth label $y_t \in \bbY$. The goal is to learn a model that performs well on future examples while keeping the number of queries at minimum to avoid the burden of the user. Label noise occurs when the user provides a label $\noisyy$ that is wrong, i.e., $\noisyy \neq y_t$. This may be caused by inattention, failing to understand the question or reporting a socially desirable label. Label noise affects the model performance \cite{frenay2014classification}. To acquire a clean dataset and a high-quality model, Skeptical Learning (\skel) was introduced by \cite{zeni2019fixing} and revised in \cite{bontempelliLearningWildIncremental2021}. 

\begin{wrapfigure}[12]{r}{0.5\textwidth}
\vspace{-4.5em}
\begin{minipage}{0.5\textwidth}
\begin{algorithm}[H]
\begin{algorithmic}[1]
    \For{$t = 1, 2, \ldots$}
        \State receive $\vx_t$ 
        \State predict $\predy_t$ for $\vx_t$ 
        \If{uncertain about $\predy_t$}
            \State request label, receive $\noisyy_t$ 
            \If{skeptical about $\noisyy$}
                \State challenge user with $\predy_t$, receive $y'_t$ 
            \EndIf
            \State add $(\vx_t, y'_t)$ to data set
            \State update classifier
        \EndIf
    \EndFor
\end{algorithmic}
\caption{Pseudo-code of \skel.}
\label{fig:skel}
\end{algorithm}
\end{minipage}
\end{wrapfigure}

\cref{fig:skel} outlines the pseudo-code of \skel. In brief, \skel sequentially learns from incoming examples. In each iteration, the machine receives an example $\vx_t$ for which predicts $y_t$. If the machine is not sufficiently confident about the prediction, it requests the user to label the uncertain example, as it is more likely to impact the model. Once the label is received, \skel has to decide whether to challenge the user's label. A label is suspicious if the model is confident that the model prediction is right and the user’s annotation is wrong. Thus, the model estimates on its own prediction and on the user's annotation.
Then, the user is asked to revise his/her annotation if the prediction has higher confidence than the annotation. 
The confidence computation depends on the actual implementation. \cite{zeni2019fixing} empirically estimates the accuracy of the machine based on training set size and confidence reported by the model. The confidence in the user is derived from the number of past spotted mistakes. \cite{bontempelliLearningWildIncremental2021} estimates the difference between the model's uncertainty on both the annotator label and predicted label. In this user study, we focus on the latter.

The first formulation of \skel has been implemented on top of random forests (RF) \cite{zeni2019fixing}. While RF is robust to noise, this method tends to be over-confident, thus avoiding querying the user on new and informative examples or continuously contradicting the user regardless of his/her past performance. Another limitation is the difficulty of fine-tuning its parameters in the interactive setting. To overcome these limitations in interactive classification in the wild, a redesign of \skel based on Gaussian Processes (GPs), non-parametric distributions over functions, has been proposed~\cite{bontempelliLearningWildIncremental2021}. GPs are defined by a mean and covariance function, where the latter encodes the assumptions about the modeled functions. The uncertainty is estimated by the prior assumption about the function and the observed data. In a nutshell, uncertainty decreases close to the observed training examples.
\skel leverages the explicit model uncertainty estimation of GPs to decide when to query, overcoming the pathological cases of the previous \skel formulation and better allocating the labeling and contradiction queries. Moreover, it supports efficient incremental learning model updates, making it suitable for the settings of this work. Thus, for this study, we used \skel on GPs and operationalized it as described in the next section.

\subsection{\skel in the wild}\label{sec:adapt}

The goal is to use \skel to assist users in answering the contextual questions and to improve answer quality. For each user, the \skel model is trained on the sensor data and annotations that arrive as a stream of data. The annotations are given by the user and capture his/her subjective view. If \skel is suspicious about the provided annotation, then the user can revise it if necessary. Acquiring correct annotations is crucial to learn an effective model that can assist the user.
The algorithm, as described in the previous section, has been slightly adapted and organised into two phases, as follows.

There is an initial bootstrap phase in which the algorithm collects annotations and trusts them regardless of the machine suspicion. The duration of this phase is fixed and addresses the problem of the cold start, which occurs when there is not enough data for a specific user.
The drawback of not being suspicious about any annotations is that the model may start learning from noisy examples. This can be addressed by reporting the past examples as an explanation supporting the machine's suspicion~\cite{tesoInteractiveLabelCleaning2021}.
In this work, we want to stress that the user is in control of the data cleaning, namely, he/she is asked by the machine to revise his/her own data by providing the annotation reflecting her/his personal point of view. 

Hence, in the second phase, the user collects his/her data and makes it fit the purpose of the data collection.
The previous formulation of \skel asked to revise previous labels as soon as the machine receives the example and decides to be skeptical. This implies that skeptical questions can continuously interrupt the user, increasing the number of interruptions. To avoid this, the skeptical questions generated during the day are sent all together at once. Potentially, this approach could allow the user to answer them in an aggregate manner, for instance,  by selecting on a map the location where they spent the morning, without having to annotate all the examples collected during that period. 
  
\section{Study Design}
\label{sec:exp:design}

This section presents the research protocol of the \skel evaluation study, which involved a multi-disciplinary team composed of a sociologist and software engineers. We employ \skel to recognize the location of the participants.
The experiment focuses only on the spatial dimension for three main reasons.
%
First, the location dimension is easier to recognize by the machine with respect to the other dimensions, like activity recognition or social context.
Second, taking into account all context dimensions would require the user to answer more questions, one for each dimension, increasing the user effort and making it more difficult to evaluate \skel.
Finally, it is possible to compute the ground-truth position from the GPS coordinates of the University of Trento and of the main home. The ground-truth labels can then be compared with the labels provided by the user.
To run the study, we integrated \skel in an existing data collection platform~\cite{kayongoMethodologyPlatformHighquality2025}.\footnote{We refer the reader to \cite[Section 7.3]{bontempelli2024human} for a detailed overview of the architecture implementing this integration.}

\subsection{Research Protocol}

\begin{figure}[tb]
    \centering
    \begin{tikzpicture}

\tikzstyle{phase} = [rectangle, rounded corners, minimum width=3cm, text width=3cm, minimum height=1.5cm, text centered, draw=black, fill=blue!10]

\tikzstyle{desc} = [rectangle, minimum width=3cm, minimum height=2cm,text centered, draw=black, text width=3cm]

\tikzstyle{week} = [rectangle, minimum width=1cm, minimum height=1cm,text centered, draw=black, text width=1.3cm]

\node (intake1) [phase] {Intake};
\node (skel1) [phase, right of=intake1, xshift=2.5cm] {Part 1:\\Bootstrapping ML model};
\node (skel2) [phase, right of=skel1, xshift=2.5cm] {Part 2:\\refining ML model};
\node (skel3) [phase, right of=skel2, xshift=2.5cm] {Part 3:\\ prediction evaluation};
\node (intake2) [phase, right of=skel3, xshift=2.5cm] {End};

\node (intake1desc) [desc, below of=intake1, yshift=-1.0cm] {Invitation};
\node (skel1desc) [desc, right of=intake1desc, xshift=2.5cm]{Questions\\Sensors};
\node (skel2desc) [desc, right of=skel1desc, xshift=2.5cm]{Questions\\Skeptical questions\\Sensors};
\node (skel3desc) [desc, right of=skel2desc, xshift=2.5cm]{Evaluation questions\\Sensors};
\node (intake2desc) [desc, right of=skel3desc, xshift=2.5cm]{Questionnaire
};

\node (1) [week, above of=intake1, yshift=0.40cm] {1st week};
\node (2) [week, above of=skel1, yshift=0.4cm] {2nd week};
\node (3) [week, above of=skel2, yshift=0.4cm, xshift=-0.87cm] {3rd week};
\node (4) [week, right of=3, xshift=+0.7cm] {4th week};
\node (5) [week, above of=skel3, yshift=0.4cm] {5th week};
\node (6) [week, above of=intake2, yshift=0.4cm] {6th week};

\end{tikzpicture}%
    \caption[Research protocol of the real-world experiment]{Research protocol. Blue boxes report the study phase and the white boxes the instruments.}
    \label{fig:exp:protocol}
\end{figure}
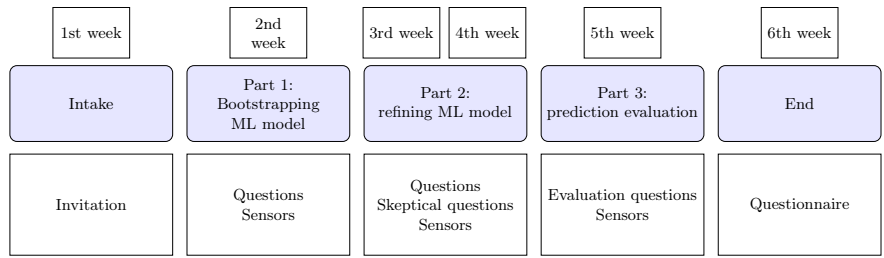

\cref{fig:exp:protocol} outlines the phases of the research protocol. The overall length of the experiment is six weeks, and four of them are allocated to the data collection.

\paragraph{Intake.}
We contacted the students of the University of Trento via email to present the research project and to provide instructions on how to join the study by installing the \ilog app \cite{zeni2014multi,kayongoMethodologyPlatformHighquality2025} on Android devices. The installation on the personal device allows the collection of data that faithfully describes their daily life and does not alter their daily routines.  Participating students must read and accept the privacy information through the app and authorize data collection for each sensor individually. The incentive strategy includes bonuses as follows: \EURcr 30 to all participants with at least 75\% completed questions, prizes of \EURcr 100 to three randomly selected most active participants.

\begin{figure}[tb]
    \centering
    \begin{subfigure}[b]{0.3\textwidth}
         \centering
         \includegraphics[width=\textwidth]{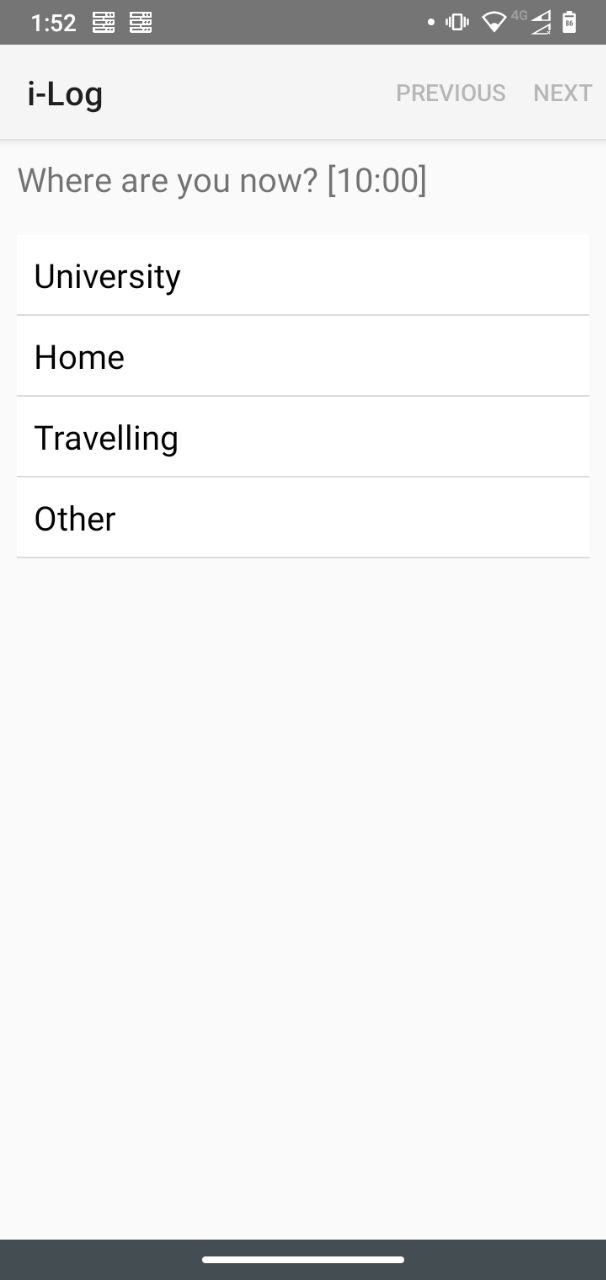}
         \caption{Time diary answer options (Q1)}
         \label{fig:screen:timediary}
     \end{subfigure}
     \hfill
     \begin{subfigure}[b]{0.3\textwidth}
         \centering
         \includegraphics[width=\textwidth]{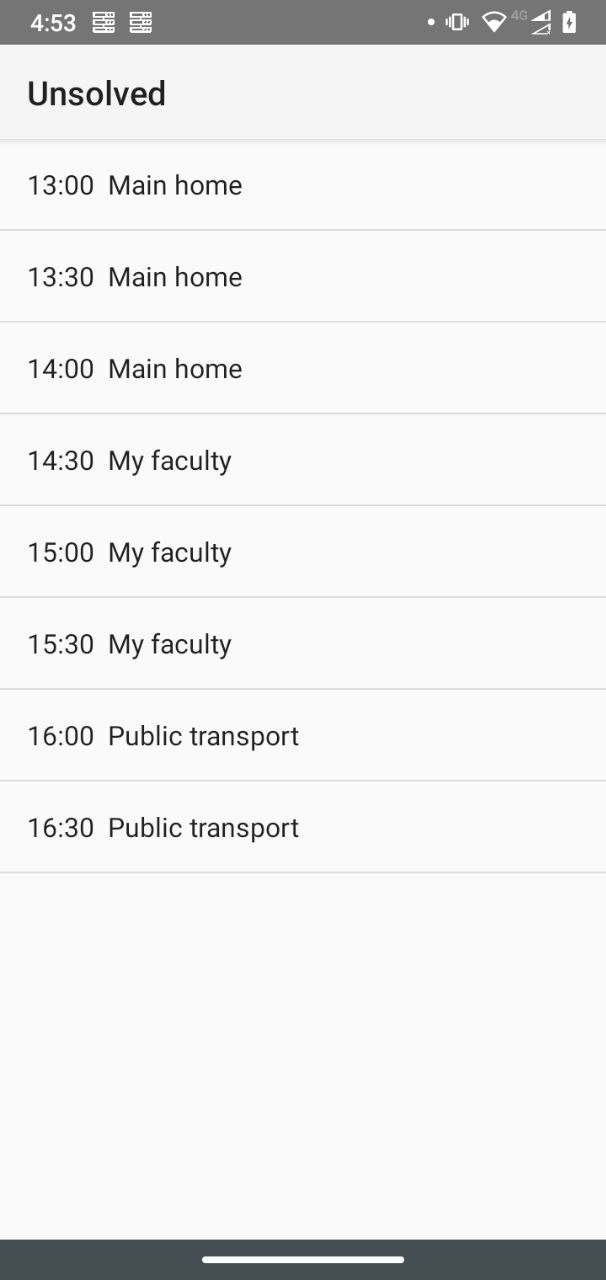}
         \caption{List of skeptical questions to revise (Q2)}
         \label{fig:screen:skel}
     \end{subfigure}
     \hfill
     \begin{subfigure}[b]{0.3\textwidth}
         \centering
         \includegraphics[width=\textwidth]{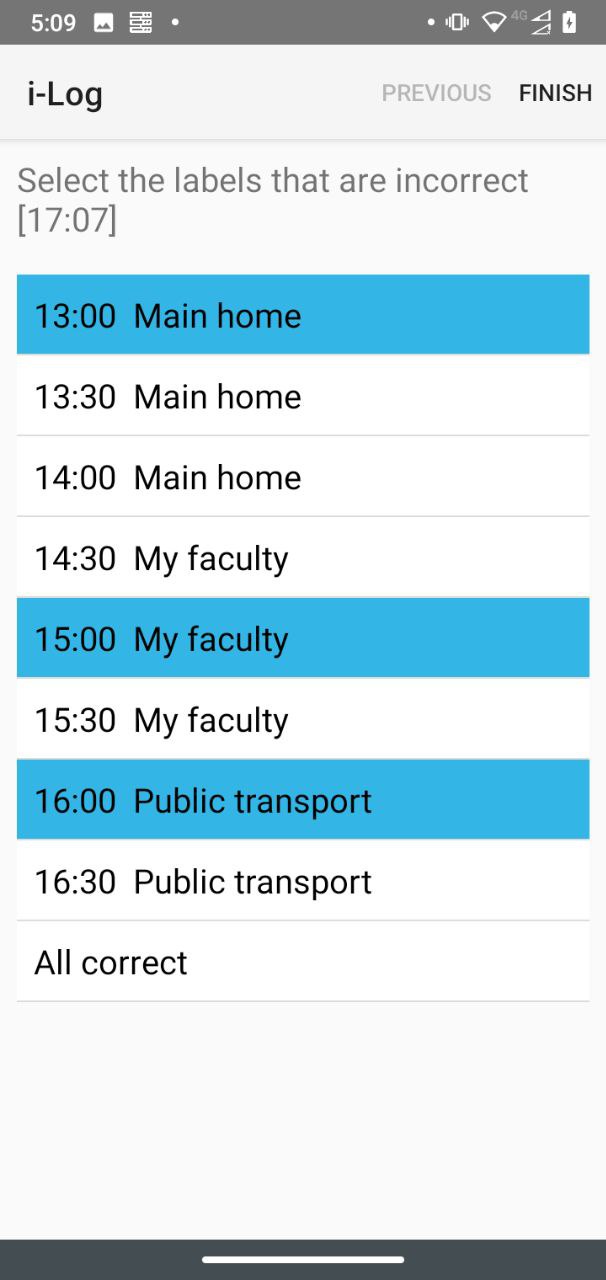}
         \caption{List of annotations to evaluate (Q4)}
         \label{fig:screen:eval}
     \end{subfigure}
    \caption[\ilog app screenshots]{The three types of questions shown on the \ilog app.}
    \label{fig:ilogScreenshoots}
\end{figure}

\paragraph{Part 1.}
During this first data collection phase, the app collects data from the sensors for one week. Moreover, every 30 minutes, the app asks the participant's location, which is the context dimension we investigate (\cref{fig:screen:timediary} shows a screenshot of the question page in the app). \cref{tab:exp:timediary} lists all answer options. \skel algorithm bootstraps the learning on this data before transitioning to the next phase.
    
\paragraph{Part 2.}
For two weeks, the participant continues to answer the time diaries, and in addition, the model challenges the participant on suspicious labels. A label is suspicious if it is different from the predicted labels and the model is sufficiently confident that its label is correct~\cite{bontempelliLearningWildIncremental2021}. The participant can confirm the predicted labels, and in this case, the model correctly detected that the label provided as the answer to the time diary was not correct. If this is not the case, a new answer is provided. The model is then refined by considering this participant feedback. The model sends the contradictions every evening at 7 pm, all at once, to concentrate the answering effort in single and specific periods of time, and reduce interruptions. \cref{fig:screen:skel} shows an example of the contradiction question. 

\paragraph{Part 3.}
The goal of the last week of the data collection is to evaluate the model predictions and, thus, the machine-participant alignment. The participant selects the incorrect prediction location labels from a list (e.g., \cref{fig:screen:eval}). This question is the only interaction with the participant in this phase, and it occurs at 7 pm.

\paragraph{Conclusion.} In the last phase, the participant is expected to fill out a questionnaire to collect socio-demographic information.

\subsection{Study Setup}

\begin{table}[tb]
    \centering
     \caption[Experiment questions]{Structure of the questions. Q1 is the time diary, Q2 and Q3 are the skeptical questions, and Q4 is the evaluation question.}
    \label{tab:exp:questions}
    \begin{tabular}{p{0.05\linewidth}p{0.25\linewidth}|p{0.3\linewidth}|p{0.3\linewidth}}
    \toprule
     & 
    \textbf{Timing/ Condition} &
    \textbf{Question} &
    \textbf{Answer options} \\
    \midrule
   Q1   &
   1st week: every 30 minutes &
   Where are you now? & 
   see \cref{tab:exp:timediary}
   \\
    \midrule
    Q2  &
   2nd and 3rd week at 7:00 pm &
   Is $<$\texttt{time}$>$ $<$\texttt{predicted label}$>$ correct? &
   \begin{tabenum}
       \item Yes
       \item No
   \end{tabenum}\\
   \midrule
   Q3 &
   if Q2 = No &
   Where are you at $<$\texttt{time}$>$? &
   go to Q1 \\
   \midrule
   Q4 &
   4th week at 7:00 pm &
   Select the labels that are incorrect &
   \begin{tabenum}
       \item $<$\texttt{time}$>$ $<$\texttt{predicted label}$>$
       \item $<$\texttt{time}$>$ $<$\texttt{predicted label}$>$
       \item \ldots
       \item All correct
   \end{tabenum}\\
\bottomrule
\end{tabular}
\end{table}

\begin{table}[tb]
    \caption[Time diary answer options]{List of answer options to the time diary question \textit{Where are you now?}.}
    \label{tab:exp:timediary}
    \centering
    \begin{tabular}{lp{0.50\linewidth}}
    \toprule
    \textbf{Main category} & \textbf{Subcategory} \\
    \midrule
    University &
    \begin{tabenum}
       \item My faculty
       \item Other faculty (UniTn)
       \item Other
   \end{tabenum}
   \\
   \midrule
    Home &
    \begin{tabenum}
       \item Main home
       \item Weekend home or holiday apartment
       \item Other people's home
   \end{tabenum}
   \\
   \midrule
   Travelling &
   \begin{tabenum}
       \item Foot
       \item Bicycle
       \item Moped, motorcycle or motorboat
       \item Passenger car
       \item Other private transport mode
       \item Public transport
   \end{tabenum}
   \\
   \midrule
   Other &
    \begin{tabenum}
       \item Restaurant, cafe, or pub
       \item Shopping centers, malls, market, other shops
       \item Hotel, guesthouse, camping site
       \item Street, square, city park
       \item Sports center
       \item Other
   \end{tabenum}
   \\
\bottomrule
\end{tabular}

\end{table}

\paragraph{Sensors.} The sensor readings are continuously collected during the four weeks of data collection. The full list of the collected sensors is presented in \cref{tab:sensor}, which also reports their collection frequency. In this experiment, we use a subset of the 30 sensors supported by the application to avoid draining the battery excessively. We selected the sensors that are more informative in predicting the location. The data streams are temporarily stored on the device and updated on the server periodically.
The raw sensor data are aggregated in time windows of 30 minutes, which is the time between two consecutive annotations. The generated feature vectors, described in \cref{tab:exp:feature},  are the input of the model.

\paragraph{Questions.} \cref{tab:exp:questions} lists the questions. Time diaries are sent every 30 minutes, and the user is asked to indicate his or her location. The list of location options (see \cref{tab:exp:timediary}) is derived from the guidelines for time use surveys~\cite{hetus}. To reduce the user effort, the options are aggregated into main categories. Skeptical questions are sent once a day, one for each suspicious answer. One evaluation question is sent daily in the final phase, listing the predicted labels, from which the user must select the incorrect ones. If time diaries and questions are not answered within 8 and 12 hours, respectively, then they expire and cannot be answered.

\paragraph{Hyperparamters.} We employ the  \skel version proposed by \cite{bontempelliLearningWildIncremental2021}. This learner leverages Gaussian Process (GP) and we modified it as described in \cref{sec:adapt}. All GP-methods use a combination of constant (with a constant value of 1), rational quadratic (with a length scale of 0.2 and a scale mixture parameter of 1), squared exponential (length scale of 1) and white noise kernels, and $\rho = 10^{-8}$, without any optimization.

\section{Results}
\label{sec:exp:analysis}

This section presents the main results and statistics about the sensor data, interaction with the participants and performance of \skel.
The number of participants that downloaded and installed the \ilog application on their devices is 77, 58 uploaded sensor data and answers. During the data collection, we sent a questionnaire to collect demographic information. We obtained the data from 42 participants, of which 90\% consider themselves male and 10\% female. All the participants belong to the Department of Information Engineering and Computer Science of the University of Trento. Most of the participants are pursuing a bachelor's degree (74\%), and the remaining a master's degree (26\%).

\subsection{Sensor Data}

\begin{figure}
    \centering
    \includegraphics[scale=0.40]{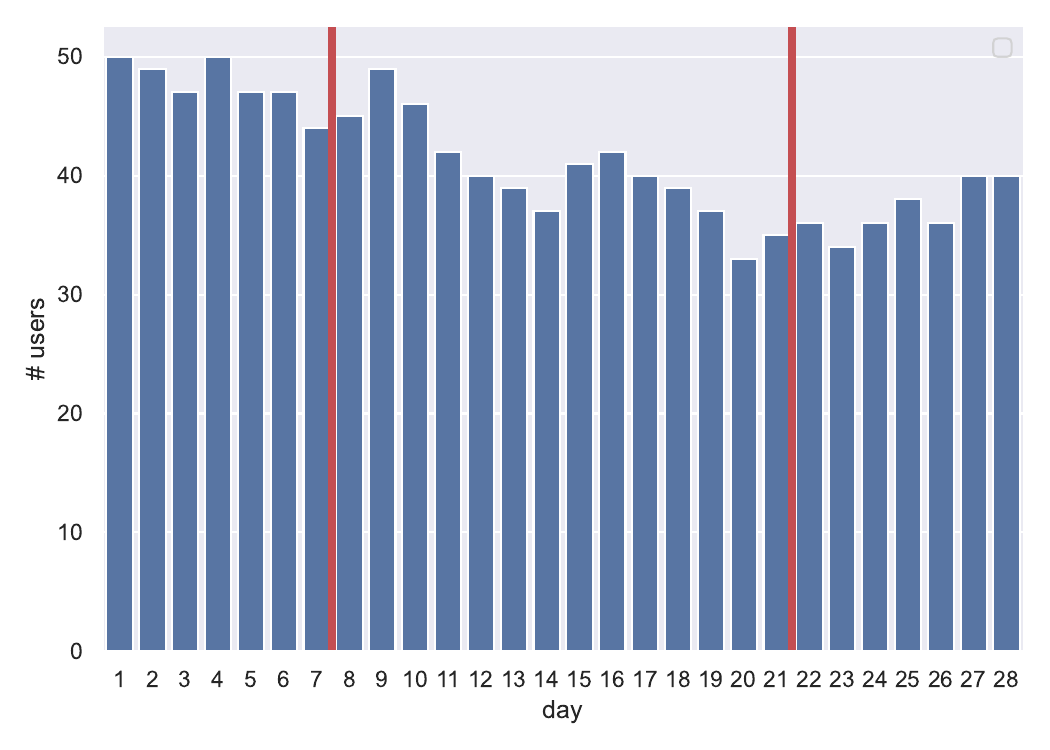}
    \caption[Number of users by day]{Number of users who uploaded sensor data by day of the experiment. Red lines divide the three data collection phases.}
    \label{fig:exp:users}
\end{figure}

The attrition effects are clearly visible in \cref{fig:exp:users} and led to participants leaving the study.  During the first week, the server received sensor data from 48 participants, whereas it decreased to 37 in the last week.
A second common problem in real-world datasets is the missing values. In this experiment, the percentage of missing values for every numeric feature varies considerably, as shown in \cref{fig:exp:emptyValues}. The reasons for missing values can be unsupported mobile devices or participants actively disabling one or more sensors. For instance, the incompatibility of some Android versions resulted in a large number of missing values for features derived from Bluetooth.

\subsection{Time Diaries}
\label{sec:exp:tdresults}

\begin{figure}[tb]
    \centering
    \includegraphics[width=0.48\textwidth]{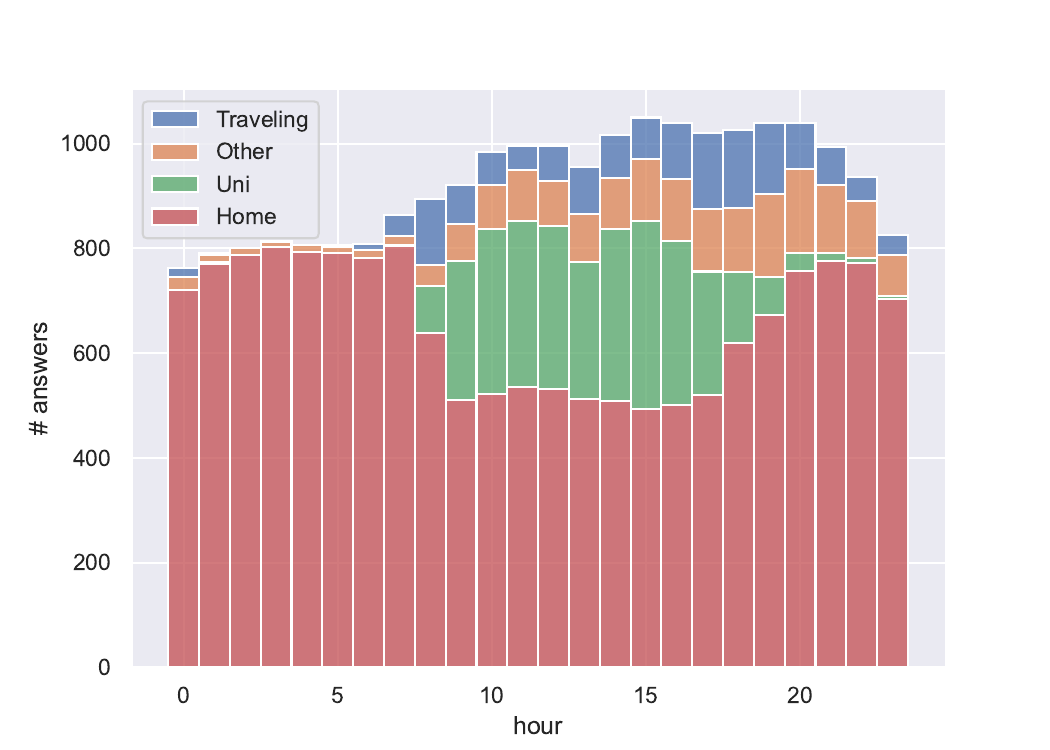}
    \includegraphics[width=0.48\textwidth]{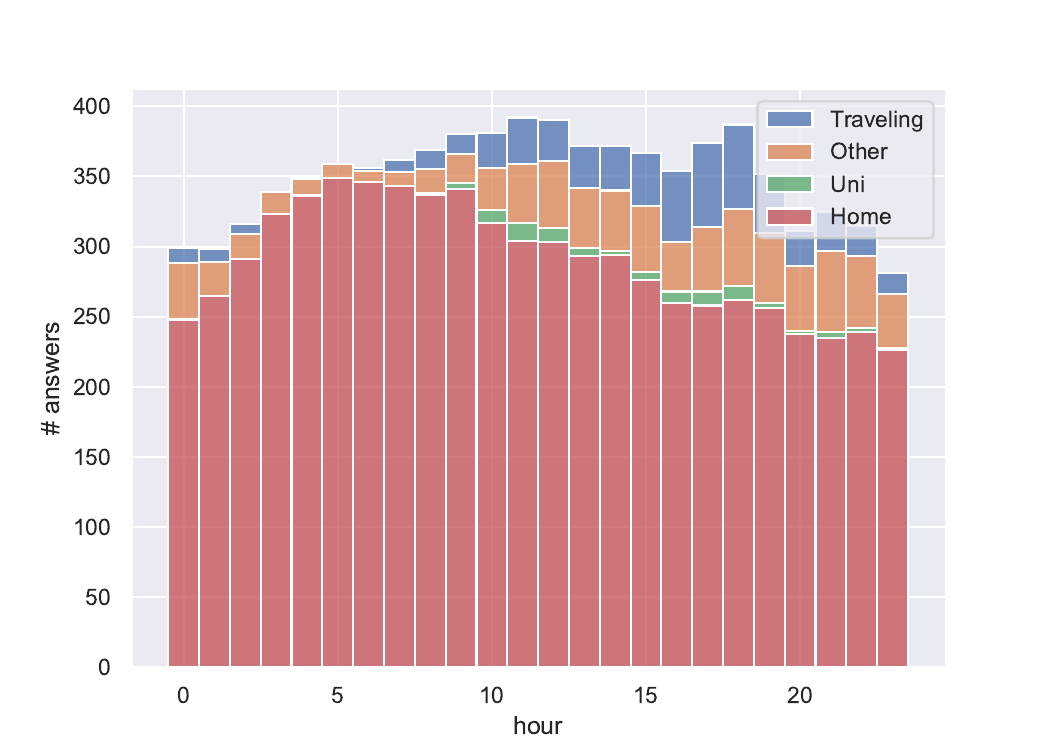}
    \caption[Number of time diary answers by hour of the day]{Number of time diary answers by hour of the day, divided by weekdays (left) and weekends (right). The answers are aggregated by main categories.}
    \label{fig:exp:labelsHours}
\end{figure}

The mobile application sends time diaries every 30 minutes. \cref{fig:exp:labelsHours} shows the distribution of the answer over hours of the day and compares weekdays and weekend days. As expected, the dataset is highly unbalanced given the high number of labels related to home (main home, weekend home, or others' home). The labels related to university are the ones that vary the most between weekdays and weekends. The dataset is highly unbalanced, and the data distribution shifts between weekdays and weekends, thus making the recognition task more challenging.

\begin{figure}[tb]
    \centering
    \includegraphics[width=\textwidth]{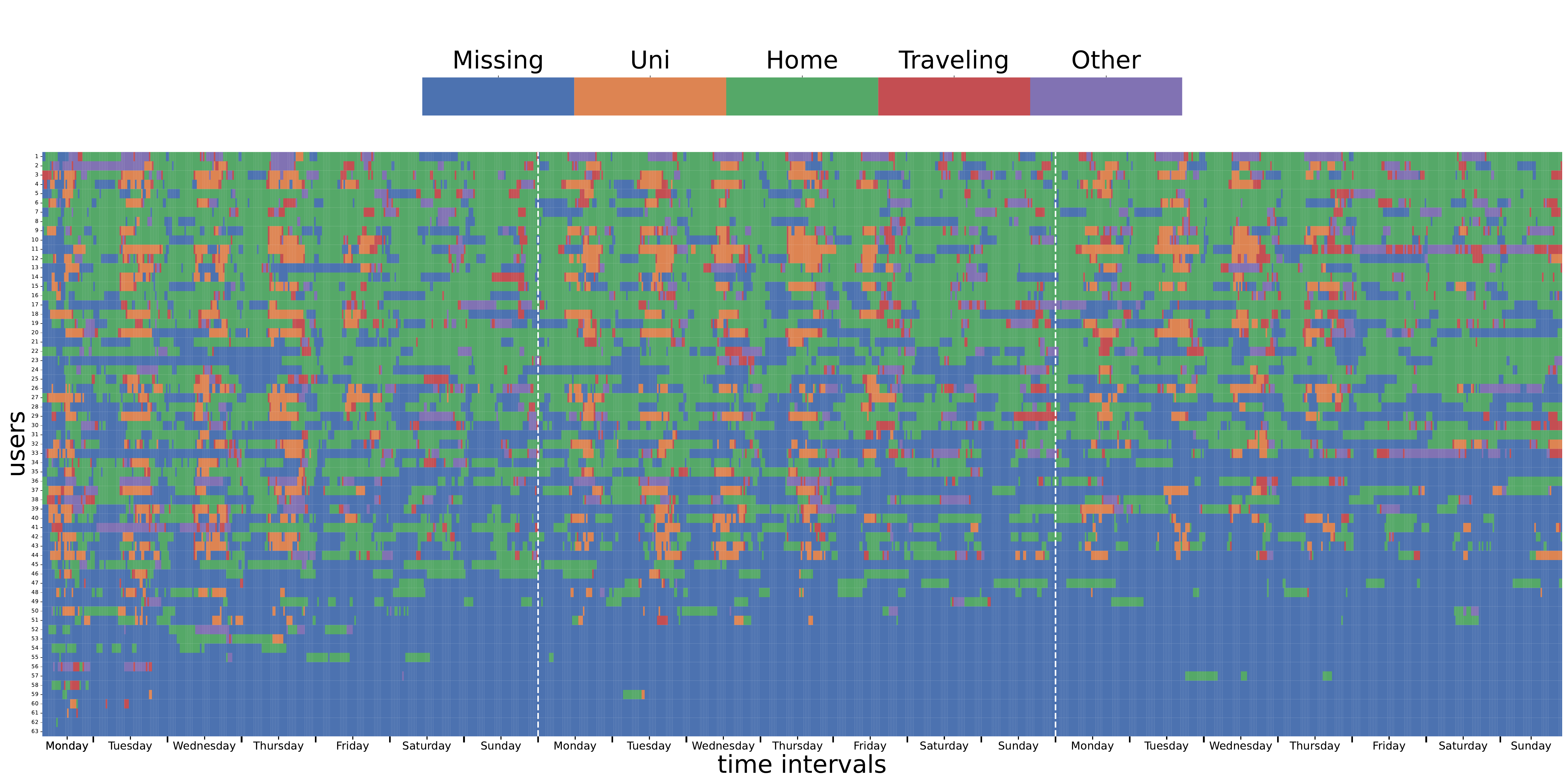}
    \caption[Main category of each label provided by the user]{The main category of the time diary answers over the first three weeks of the experiment. Rows: all users of the experiment. Columns: time interval of 30 minutes (i.e., annotation). White vertical lines denote each week.}
    \label{fig:exp:dataDiversity}
\end{figure}

\noindent
\cref{fig:exp:dataDiversity} shows the time diary answers for each user over the first three weeks. Each cell of the heatmap is an interval of 30 minutes, and the color denotes the main category of the answer, i.e., university, home, traveling and other. Blue cells are unanswered questions. Note how the answering pattern varies across users. The top rows represent users who regularly provide answers, whereas the bottom rows represent less active users, who left the experiment at a certain point. Users in the central part alternate days with answers and periods where the question expires. Thus, \skel is not helpful for all types of users, as discussed in the following sections.

\subsection{Skeptical Questions}

\begin{figure}[tb]
    \centering
    \begin{subfigure}[t]{0.48\textwidth}
        \centering
        \includegraphics[width=\linewidth]{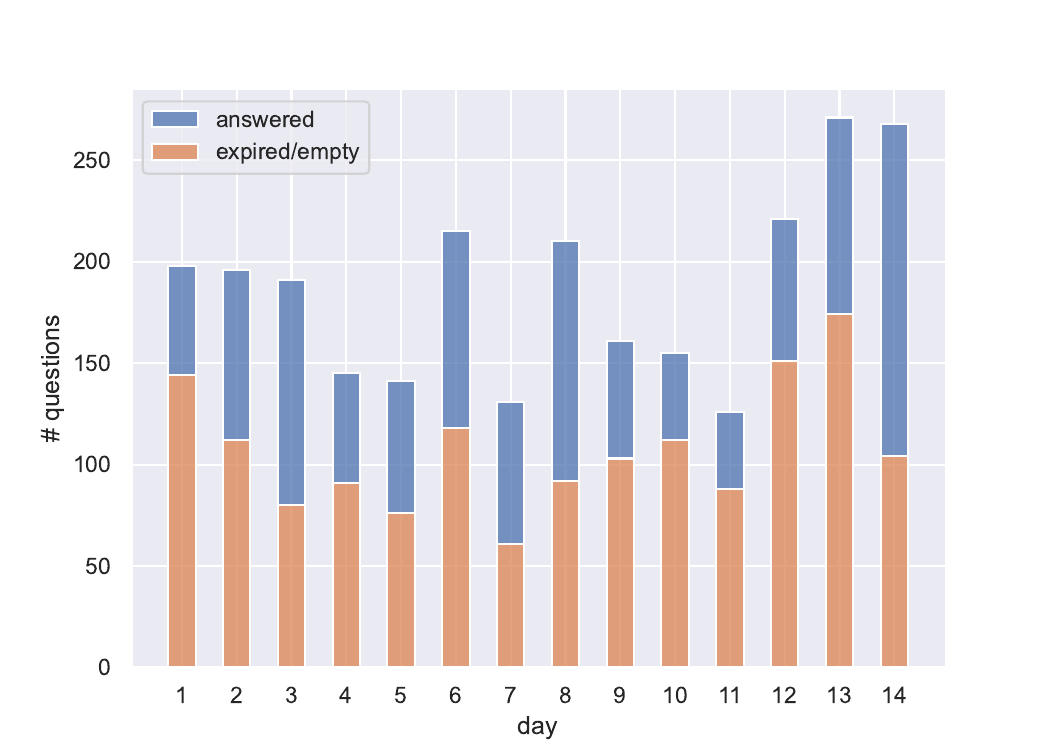}
        \caption{Number of questions with (blue) and without (orange) an answer.}
        \label{fig:exp:skelquestion:left}
    \end{subfigure}
    \hfill
    \begin{subfigure}[t]{0.48\textwidth}
        \centering
        \includegraphics[width=\linewidth]{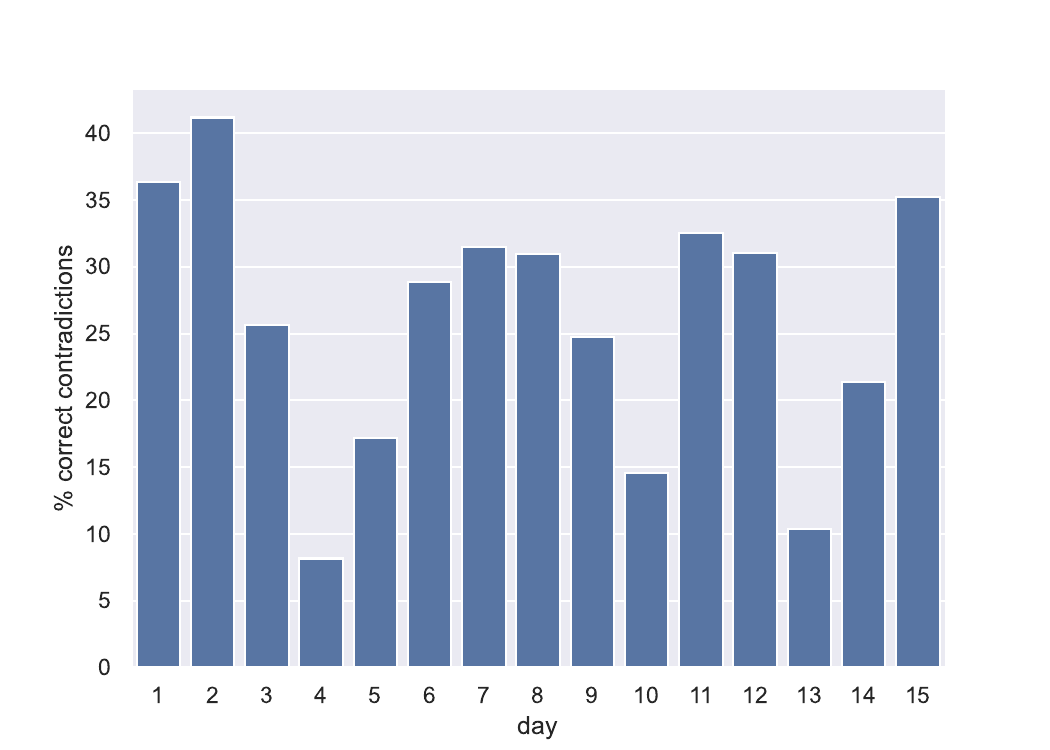}
        \caption{Percentage of skeptical contradiction in which the machine label is confirmed as correct by the user.}
        \label{fig:exp:skelquestion:right}
    \end{subfigure}
    \caption{Statistics about skeptical questions during the two weeks of Part 2 of the study.}
    \label{fig:exp:skelquestion}
\end{figure}

The time diary answers are used to train the \skel model, one model for each user. The model learns the mapping between sensor data and location labels. As described in \cite{bontempelliLearningWildIncremental2021}, mistaken labels badly affect the machine's performance. In phase two of the experiment (the second and third week of the study), the machine sends skeptical questions to contradict the user whenever it is suspicious about the label. The goal is to give the participant the opportunity to fix wrong labels.
\cref{fig:exp:skelquestion:left} shows the total number of contradictions sent to the users, split between answered and not answered. The fraction of missing answers is high, namely more than 50\% for most of the days. The main causes are that the question is not delivered because the phone was not connected to the Internet, or users did not respond in time. 
In 25\% of the answers, the machine prediction was selected as correct, whereas in the rest of the answers, the user provided a different label. When rejecting the predicted label, in 80\% of cases, the participants confirmed the label they provided the first time.  \cref{fig:exp:skelquestion:right} plots the number of times the machine was right. Therefore, when contradicted, the participants considered their answers correct most of the time, and revised their previous answers in one-fourth of the cases.

\subsection{Evaluation Questions}

\begin{figure}[tb]
    \centering
        \includegraphics[width=0.75\textwidth]{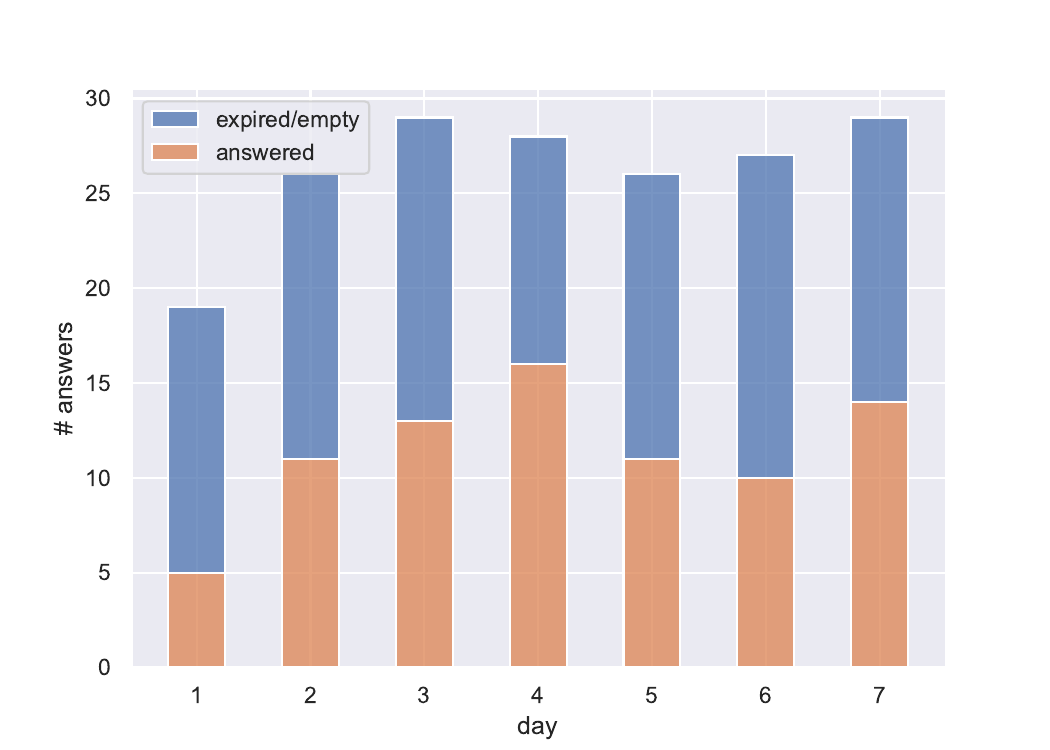}
        \caption[Statistics about the evaluation questions]{
        Total number of evaluation questions sent to the user for each day of the last week of the experiment. Orange: number of received answers.  Blue: number of questions without answer.}
        \label{fig:exp:evalQuestion}
\end{figure}
\begin{figure}[tb]
        \includegraphics[width=\textwidth]{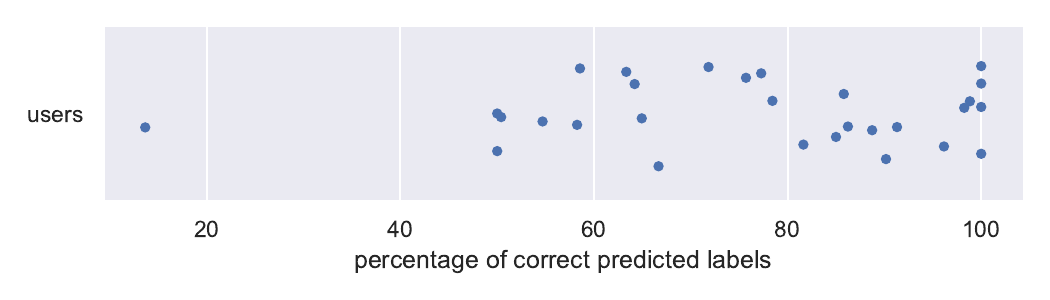}
        \caption[Correct predictions by user]{Percentage of predicted labels that are evaluated as correct by the user. Each point is a user.}
        \label{fig:exp:evalQuestionCorrect}
\end{figure}

In the third and last phase, each day, the participant is asked to evaluate the predictions of the machine. The questions, sent at 7 pm, list the location labels predicted in the last 24 hours. Then, the user selects those labels that she/he considers wrong. \cref{fig:exp:evalQuestion} shows the number of questions sent to the user for each day of the last week of the experiment. The fraction of unanswered questions is more than 50\% (blue bar). In 25\% of the received answers, the participants evaluated the prediction of that day as all correct. Each day, participants have to evaluate 30 predicted labels on average (out of 48 possible prediction in the last 24 hours).
\cref{fig:exp:evalQuestionCorrect} details, for each user, the percentage of the correct labels. The average percentage is 76\%, showing that the participants rated as correct the majority of the predicted locations, thus highlighting the potential of \skel to reduce the answering effort.

\subsection{\skel Performance}

\begin{figure}[tb]
    \centering
    \includegraphics[scale=0.55]{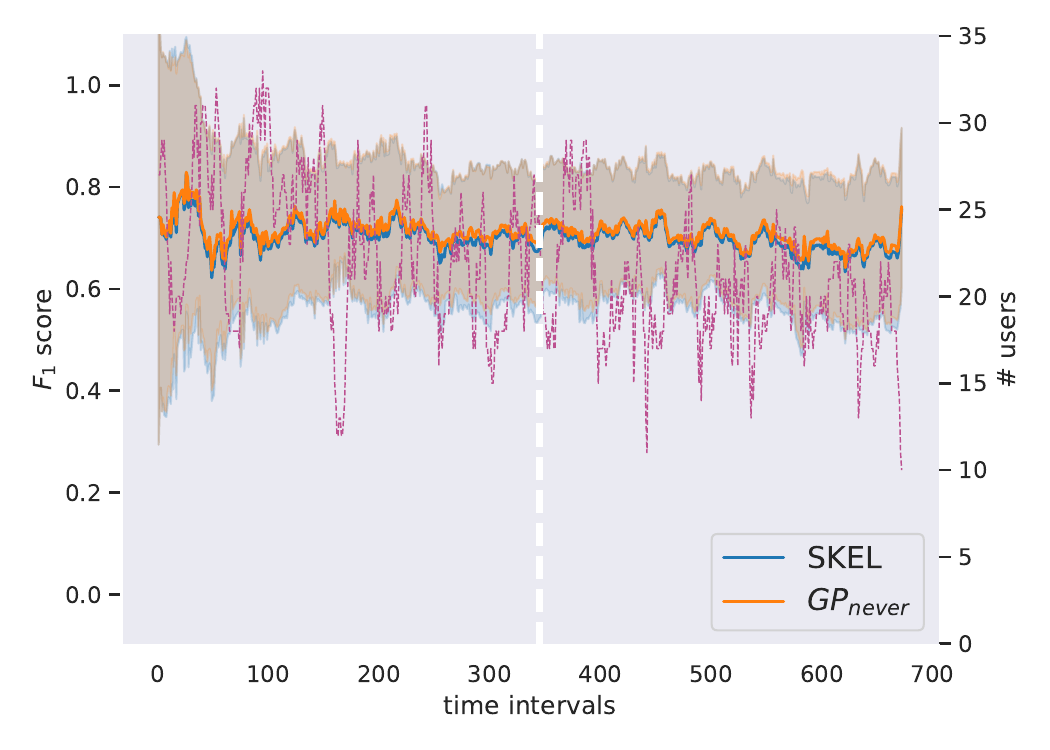}
    \caption[Progressive average $F1$-score]{Progressive $F1$-score averaged over all users. Time intervals are the 30-minute windows of two weeks. Shaded area is the standard deviation. Violet dashed line represents the number of users. Blue line is the \skel, as presented in \cite{bontempelliLearningWildIncremental2021}. Orange line: \skel variant in which the user is never contradicted.}
    \label{fig:exp:f1score}
\end{figure}

We compare the effectiveness of \skel with a variant that never contradicts the user (denoted with \gpnever). \cref{fig:exp:f1score} reports the experimental results on these two methods averaged over all users. The plot shows the progressive \fone over the third and fourth week for every 30-minute timeslot. Since the user label, which is used as ground truth, or the input data are not always available, the number of users varies over the time intervals (violet dashed line).
The performance of \skel and \gpnever are overlapping, which shows that, on average, there was no advantage in being skeptical. Indeed, in only 25\% of the contradictions, the machine was correctly skeptical about the participant supervision (see \cref{fig:exp:skelquestion:right}).
The motivations are that the participants were consistent in providing the annotation even after being contradicted. As shown in~\cite{zeni2019fixing} and discussed in the next section, the performance of \skel depends on the user behaviors. Another cause is the high fraction of missing values that impacts the predictive power of the model and thus reduces the benefits of \skel.

\subsection{Discussion}
\label{sec:exp:conclusioni}

The results show that in this study, the participants were mostly consistent with the provided supervision. This might be explained by the limited respondent burden due to the short experiment period and the focus on only one context question, i.e., the participant's location. Additionally, \cite{zeni2019fixing} identified four different prototypical users and showed how the performance of the \skel is affected by the user behaviors. Inattentive and predictable users are cases in which \skel generates substantial benefits. For the reliable user, \skel does not provide improvements, but at the same time, it is not harmful. The last type is the tricky user, for whom \skel fails to learn. However, the positive result is that the participants rated as correct the majority of the labels predicted in the evaluation week, confirming the benefits of employing the model in longitudinal studies. The benefits of skeptical learning become more visible when running longer data collections with multiple questions. Additionally, the low response rate to the questions and the missing sensor readings impacted the performance of \skel. Thus, this result highlights the difficulties of running studies in the wild with real users.

\noindent
This work, however, has some limitations. The number of participants is small and consists of only students from one university department. With respect to the \skel version presented in \cite{bontempelliLearningWildIncremental2021}, we removed the active queries because the examples were not available in real-time to the machine learning model to decide when to query, and thus, during the first three weeks, labels were asked on all incoming examples regardless of the machine prediction confidence. The evaluation questions were sent in the last week of the experiment and are designed to allow the user to select wrongly predicted labels. Participants may have selected only a subset of the mistakes to avoid going through the full list of predictions. Regarding skeptical questions, participants may have felt frustrated by the contradiction and thus consistently rejected the machine label. Future work should investigate the psychological and behavioral implications that arise from interacting with a machine that learns about your routines and contradicts your answers.

Furthermore, future experiments should consider other factors that impact the quality of the answers, such as user reaction time and completion time, time of the day, mood and situational context of the participant~\cite{bison2024impactsqualityuseranswers}. A better question scheduling to improve response rate and answer quality ~\cite{zhao2023scheduling}.
Additionally, the usability and effectiveness of the \ilog app need to be assessed to evaluate the impact on the label quality. One possible enhancement is the visualization of recognized locations on an interactive timeline, which can help users identify errors more intuitively.
The model performance can be improved by implementing per-user hyperparameter tuning, which adapts the model to the specific user. To this end, recent works proposed solutions for hyperparameter tuning on data streams~\cite{veloso2021hyperparameter,carnein2020confstream} and AutoML for online learning~\cite{celik2023online}.
Future evaluations should also explore the emotional responses of participants when interacting with the machine learning model, especially when being challenged by the machine. The future work includes applying \skel to other domains than context recognition and extending it to multi-modal inputs.

\section{Conclusion}\label{sec:conclusion}

We designed and executed a study with real users in the wild to evaluate \skel. The study focused on a social science use case,   specifically, longitudinal studies in which participants provide information about their daily lives multiple times per day. In this context, the results show the potential of \skel to reduce respondent burden by allowing the system to automatically answer questions when it is sufficiently confident in its predictions. The number of questions can thus be reduced, allowing the researcher to increase the duration of data collection and mitigating the drop-out problem. Moreover, \skel ensures the quality of the collected labels by involving the users in fixing their answers.

\begin{credits}
\subsubsection{\ackname}
The research leading to these results has received funding from the European Union’s Horizon 2020 Research and Innovation Programme, through the TRUMAN project, under Grant Agreement No. 101214000. Views and opinions expressed are, however, those of author(s) only and do not necessarily reflect those of the European Union or European Health and Digital Executive Agency (HADEA). Neither the European Union nor the granting authority can be held responsible for them.
The content in \cref{sec:exp:design,sec:exp:analysis} is based on Chapter 7 of AB's PhD thesis~\cite{bontempelli2024human}.
We acknowledge the use of Grammarly as a tool for grammar refinement.

\subsubsection{\discintname}
The authors have no competing interests.
\end{credits}
%
%
%
\bibliographystyle{splncs04}
\bibliography{bib/exp,bib/main,bib/skel}

\appendix
\section{Appendix}

\begin{figure}
    \centering
    \includegraphics[scale=0.45]{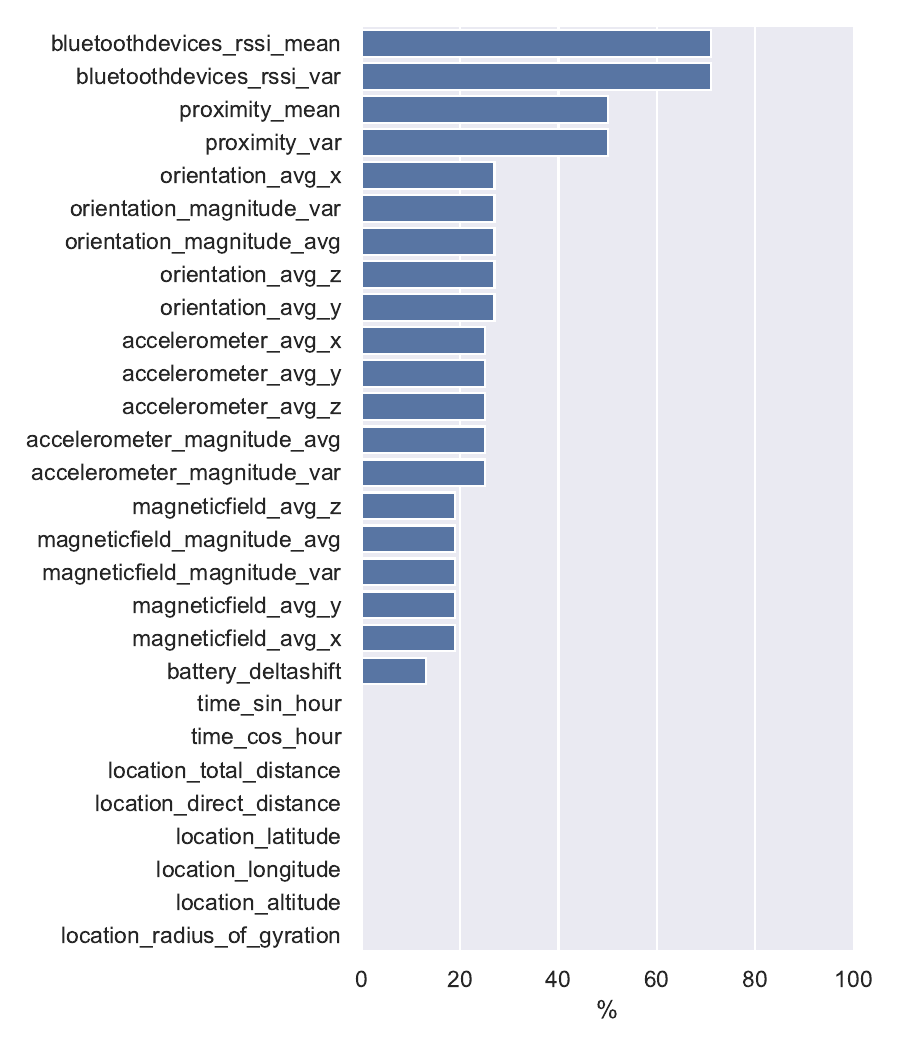}
    \caption{Percentage of missing values for each numeric feature.}
    \label{fig:exp:emptyValues}
\end{figure}

\cref{tab:sensor} provides an overview of the sensor data collected in the study, and \cref{tab:exp:feature} details the engineered features utilized as inputs to the model. The distribution of missing values is illustrated in \cref{fig:exp:emptyValues}.

\begin{center}
\begin{longtable}{>{\raggedright\arraybackslash}p{0.25\linewidth}|p{0.71\linewidth}}
    \caption[\ilog sensors]{List of \ilog sensors collected during the four weeks of experiment.} \label{tab:sensor} \\
    \toprule
    \textbf{Sensor} & \textbf{Description} \\
    \midrule
    \endfirsthead
    \multicolumn{2}{l}{\textbf{\textsc{Connectivity}}} \\
    Bluetooth normal, Bluetooth low energy & 
    Returns the discovered Bluetooth normal or low energy devices.\\
    WiFi Event & 
    Returns information related to the WIFI network to which the phone is connected; if connected, it also reports the WIFI network ID.\\
    WiFi Networks Event &
    Returns all WIFI networks detected by the smartphone.\\
    \multicolumn{2}{l}{\textbf{\textsc{Activity}}} \\
    Accelerometer & 
    Returns the acceleration of the device along the three coordinate axes.\\
    Activities &
    Return the user's activity recognized by the Google Activity Recognition API. The recognized activities are \textit{in vehicle}, \textit{on bicycle}, \textit{on foot}, \textit{running}, \textit{still}, \textit{tilting}, \textit{walking} and \textit{unknown}. The sensor reports a confidence score between 0 and 100, which represents the likelihood that the user is performing the activity.\\
    Step detector & 
    An event is triggered each time the user takes a step.\\
    Orientation & 
    Returns the position of the device relative to the earth's magnetic north pole.\\
    \multicolumn{2}{l}{\textbf{\textsc{Location}}} \\
    Location event & GPS coordinates (latitude, longitude and altitude) \\
    Magnetic field &
    Reports the ambient magnetic field along the three sensor axes.\\
    Proximity Event & Measures the distance between the user's head and the phone. Depending on the phone, it may be measured in centimetres (i.e., the absolute distance) or as labels (e.g., 'near', 'far')\\
    \multicolumn{2}{l}{\textbf{\textsc{Software}}} \\
    Battery Charge Event &
    Returns whether the phone is on charge and the type of charger\\
    Battery Monitoring Log & 
    Returns the phone's battery level\\
    \bottomrule
\end{longtable}
\end{center}
\begin{center}
    \begin{longtable}{p{0.4\linewidth}p{0.1\linewidth}p{0.5\linewidth}}
        \caption[Engineered features]{List of features generated by aggregating raw sensor data in windows of 30 minutes.} \label{tab:exp:feature} \\
        \toprule
        \textbf{Feature name} & 
        \textbf{Type} & 
        \textbf{Description}  \\
        \midrule
        \endfirsthead
        \multicolumn{3}{l}{\textbf{\textsc{Time}}} \\
        \texttt{time\_is\_workday} & 
        boolean &
        True for the days from Monday to Friday\\
        \texttt{time\_is\_morning} & 
        boolean &
        True for the hours between 6 am and 9 am\\
        \texttt{time\_is\_noon} & 
        boolean &
        True for the hours between 10 am and 1 pm\\
        \texttt{time\_is\_afternoon} & 
        boolean &
        True for the hours between 2 pm and 5 pm\\
        \texttt{time\_is\_evening} & 
        boolean &
        True for the hours between 6 pm and 9 pm\\
        \texttt{time\_is\_night} & 
        boolean &
        True for the hours between 10 pm and 5 am\\
        \texttt{time\_sin\_hour}, \texttt{time\_cos\_hour} &
        float &
        Sine and cosine transformations of the hour to encode a stronger connection between two nearby hours
        \\[10pt]
        \multicolumn{3}{l}{\textbf{\textsc{Connectivity}}}\\
        \texttt{bluetoothdevices\_rssi\_ \{mean,var\}} &
        float & 
        Mean and variance of the Received Signal Strength Indicator (RSSI) of the detected Bluetooth devices\\
        \texttt{bluetoothdevices\_nunique} &
        integer &
        Number of unique Bluetooth normal and low energy devices \\
        \texttt{wifi\_connection\_count} &
        integer &
        number of times the device connected to a WiFi network\\
        \texttt{wifi\_is\_connected} &
        boolean &
        True if the devices connected to a WiFi network at least once\\
        \texttt{wifinetworks\_nunique} &
        integer &
        Number of unique networks detected\\[10pt]
        \multicolumn{3}{l}{\textbf{\textsc{Activity}}}\\
        \texttt{step\_detection\_count} &
        integer &
        Number of step detection events\\
        \texttt{activity\_ \{invehicle,onbycicle,onfoot, running,still,unknown, walking\}} & 
        boolean &
        True if the Google activity recognition API has recognized the activity\\
        \texttt{accelerometer\_avg\_\{x,y,z\}} & 
        float &
        Mean of all accelerometer values for each axes separately\\
        \texttt{accelerometer\_magnitude\_ \{avg,var\}} &
        float &
        Mean and variance of the magnitude of each sensor reading\\
        \texttt{orientation\_avg\{x,y,z\}} &
        float &
        Mean of all orientation values for each axes separately\\
        \texttt{orientation\_magnitude\_ \{avg,var\}} &
        float &
        Mean and variance of the magnitude of each sensor reading\\[10pt]
        \multicolumn{3}{l}{\textbf{\textsc{Location}}}\\
        \texttt{location\_ \{altitude,longitude,latitude\}} &
        float &
        Averaged GPS coordinates\\
        \texttt{location\_direct\_distance} &
        float &
        Distance between the first and last location point\\
        \texttt{location\_total\_distance} &
        float &
        Total distance covered \cite{canzian2015trajectories}\\
        \texttt{location\_radius\_of\_gyration} & float & Deviation from the centroid of the GPS points \cite{pappalardo2013understanding,gonzalez2008understanding,yue2014zooming}\\
        \texttt{magneticfield\_avg\{x,y,z\}} &
        float &
        Mean of all magnetic field values for each axis separately\\
        \texttt{magneticfield\_magnitude\_ \{avg,var\}} &
        float &
        Mean and variance of the magnitude of each sensor reading\\
        \texttt{proximity\_\{mean,var\}} &
        float &
        Mean and variance of the proximity values\\[10pt]
        \multicolumn{3}{l}{\textbf{\textsc{Software}}} \\
        \texttt{battery\_deltashift} &
        float &
        Battery level difference between the beginning and the end of the interval\\
        \texttt{battery\_charge\_count} &
        integer &
        Number of times the phone has been connected to a charging source during the interval\\
        \bottomrule
    \end{longtable}
\end{center}

\end{document}